\documentclass{ifacconf}

\usepackage{amsfonts,latexsym,amsmath,amssymb}
\usepackage[mathscr]{eucal}
\usepackage{comment}
\usepackage{natbib}
\usepackage{graphicx,url}
\usepackage{color}
\usepackage{epstopdf}
\usepackage{mathrsfs}

\usepackage{ifthen}
\newboolean{short}
\setboolean{short}{False} 

\newtheorem{nnassumption}{\bf Assumption}

\newtheorem{nntheorem}{\bf Theorem}

\newtheorem{nndefinition}{\bf Definition}

\newtheorem{nnproposition}{\bf Proposition}

\newtheorem{nnproblem}{\bf Problem}

\newtheorem{nnlemma}{\bf Lemma}

\newtheorem{nnclaim}{\bf Claim}

\newtheorem{nnremark}{\bf Remark}

\newenvironment{proofof}{{\em Proof of}}{\hfill \hspace*{1pt}\hfill $\Box$}

\usepackage{amsmath}
\usepackage{color}

\definecolor{forestgreen}{rgb}{0.13,0.54,0.13}

\renewcommand{\epsilon}{\varepsilon}
\newcommand\cW{\mathbf{W}}
\newcommand\cL{\mathbf{L}}
\newcommand\cP{\mathbf{P}}
\newcommand\cZ{\mathbf{Z}}
\newcommand\cF{\mathbf{F}}
\newcommand\csigma{\boldsymbol{\sigma}}

\newcommand\RR{\mathbb{R}}
\newcommand\NN{\mathbb{N}}

\begin{document}

\begin{frontmatter}

\title{A Unified
Representation of \\
Neural Networks Architectures}

\author[Grenoble]{Christophe Prieur}
\author[Eindhoven]{Mircea Lazar}
\author[Grenoble]{Bogdan Robu}

\address[Grenoble]{Univ. Grenoble Alpes, CNRS, Grenoble INP \\GIPSA-lab \\Grenoble, France}
\address[Eindhoven]{Eindhoven University of Technology\\
Electrical Engineering, Control Systems       \\
       Eindhoven, The Netherlands}

\begin{abstract}
In this paper we consider the limiting case of neural networks (NNs) architectures when the number of neurons in each hidden layer and the number of hidden layers tend to infinity thus forming a continuum, and we derive approximation errors as a function of the number of neurons and/or hidden layers. Firstly, we consider the case of neural networks with a single hidden layer and we derive an  infinite width integral neural representation that generalizes existing continuous neural networks (CNNs) representations. Then we extend this to deep residual CNNs that have a finite number of integral hidden layers and  residual connections. Secondly, we revisit the relation between neural ODEs and deep residual NNs and we formalize approximation errors via discretization techniques. Then, we merge these two approaches into a unified homogeneous representation of NNs as a Distributed Parameter neural Network (DiPaNet) and we show that most of the existing finite and infinite-dimensional NNs architectures are related via homogenization/discretization with the DiPaNet representation. \ifthenelse{\boolean{short}}{}{Our approach is purely deterministic and applies to general, uniformly continuous matrix weight functions. Relations with neural fields and other neural integro-differential equations are discussed along with further possible generalizations and applications of the DiPaNet framework.}
\end{abstract}

\begin{keyword}
Neural networks, Continuous neural networks, \ifthenelse{\boolean{short}}{}{Residual neural networks, Neural ODEs,} Distributed parameter systems
\end{keyword}
\end{frontmatter}

\section{Introduction}
Deep neural networks (NNs) with multiple hidden layers for function approximation emerged in the pioneering work of \citep{ivakhnenko1967gmdh} and have dominated the landscape of artificial intelligence and deep learning every since, see, for example, the \ifthenelse{\boolean{short}}{survey \citep{lecun2015deep}}{surveys \citep{lecun2015deep, schmidhuber2015deep}}. Deep NNs typically feature a discrete representation that comprises a finite number of hidden layers (i.e., \emph{the depth}) and a finite number of neurons in each hidden layer (i.e., \emph{the width}). \ifthenelse{\boolean{short}}{}{Universal approximation results for NNs originally considered networks with a single layer, see, e.g., \citep{cybenko1989approximation}.}

One of the first papers that derived an integral form of an infinite width \ifthenelse{\boolean{short}}{NN is \citep{Barron1993Universal}.}{ (i.e., corresponding to summing up over an infinite number of neurons within a single hidden layer) NN is \citep{Barron1993Universal}, which represents functions in a Barron space as a superposition of ridge functions and integrates over the parameter space of weights and biases. The purpose of this work was to obtain tighter function approximation bounds for shallow NNs.} An alternative approach was developed by \citep{cnn}, who derived integral continuous neural networks (CNNs) with a single hidden layer by integrating over an arbitrary domain in which the neuron index takes values. Therein it was shown that finite-dimensional single layer NNs can be recovered via discretization techniques and weights represented as piecewise constant functions\ifthenelse{\boolean{short}}{.}{; while other type of functions, such as affine functions, could lead to new families of NNs corresponding to different discretization methods.} Recently, seemingly unaware of \citep{cnn}, a similar class of integral continuous neural networks has been derived in \citep{Solodskikh2020Multilayer} motivated by image learning. \ifthenelse{\boolean{short}}{}{Therein, a high-dimensional hypercube is employed to
represent the weights of one layer as a continuous surface and integral operators analogous to the conventional
discrete operators in neural networks are then employed.} A deep neural operator architecture for learning operators that are mappings between infinite-dimensional function spaces was recently developed in \citep{Kovachki2023}, by using the composition of a finite number of integral operators.

Other works, which address the limiting case of shallow NNs, include the neural-tangent-kernel \citep{Jacot2018Neural}\ifthenelse{\boolean{short}}{}{ and the mean-field approach \citep{Nguyen2020MeanFieldMultilayer}}, which, however, differ by considering randomly distributed initial weights and integrating over initial network parameters. These works yield infinite-width limits that collapse to kernel regression with a fixed Gaussian process prior.

Another branch of research considered the case of infinite depth NNs, i.e., a continuum of hidden layers and a finite number of neurons, which was formally defined as neural ODEs in \citep{neuralODE}. Other papers such as \citep{Weinan2017ProposalMLviaDS, LuZhong2018BeyondFinite} have established a relation between neural ODEs and residual NNs \citep{he2016deep} via a dynamical systems perspective. \ifthenelse{\boolean{short}}{}{These works established that similarly to discretization of dynamical systems (ODEs) with respect to time, one could show that deep residual NNs are discretizations of neural ODEs and vice versa. More recently, \citep{ruthotto2020deep} showed that a class of deep residual convolutional neural networks can be interpreted as nonlinear systems of PDEs.} Other examples of recent works that leverage relations between deep residual NNs and properties of dynamical systems are \citep{Tabuada}, which uses controllability to establish universal approximation properties and \citep{OCresnet}, which uses optimal control to train residual NNs.

Another approach to designing continuous neural networks, i.e., the neural fields approach, stems from the neuro-physiological perspective that treats neurons as entities distributed in space, see, e.g., \ifthenelse{\boolean{short}}{\citep{Cook2021NeuralFieldOverview}.}{\citep{Amari2000Methods, CoombesEtAl2014, Cook2021NeuralFieldOverview}.} \ifthenelse{\boolean{short}}{}{Neural fields can be regarded as neural PDEs where one integrates over the space of neurons.}Such neural field representations have been recently employed to develop the Neural Integro-Differential Equations (NIDEs) framework \citep{Zappala2023NeuralIntegroDifferential} for modeling dynamics with integro-differential equations that are determined by non-local operators. \ifthenelse{\boolean{short}}{}{Furthermore, delayed neural fields have been used for observer and controller design for PDEs in \citep{Brivadis2024AdaptiveObserver}. It is worth to mention that existing neural fields and neural integro-differential representations have been motivated by learning dynamics described by partial differential equations (PDEs) and especially brain modeling problems.}

Although neural operators \citep{Kovachki2023} and NIDEs \citep{Zappala2023NeuralIntegroDifferential} represent generalizations of neural ODEs \citep{neuralODE} to functions spaces, a neural network architecture that unifies integral continuous neural networks for function approximation as introduced by \citep{cnn} with neural ODEs is yet to be developed, to the best knowledge of the authors.

In this paper we analyze the limit of deep residual neural networks when the number of neurons and the number of hidden layers tend to infinity and we derive a novel homogeneous representation of NNs that unifies the CNN representation of \citep{cnn} with the neural ODE representation of \citep{neuralODE} via a distributed parameter systems approach. Our framework is fully deterministic and yields a neural integro-differential representation where the integration is over an arbitrary domain, e.g., $[0,1)$. The resulting homogeneous neural network can be regarded as a Distributed Parameter neural Network (DiPaNet) from which we can recover neural ODEs \citep{neuralODE}, Deep CNNs \citep{cnn}, and Deep Residual NNs \citep{he2016deep, Tabuada} by applying suitable \emph{discretization} techniques. In reverse, we can show that all these architectures converge to the DiPaNet representation through suitable \emph{homogenization} techniques.

\ifthenelse{\boolean{short}}{}{The established relations provide the fundament for deriving explicit links between the discretization level, i.e., number of neurons and hidden layers, and a desired approximation accuracy, potentially yielding mathematically rigorous criteria for dimensioning Deep Residual NNs, with important implications for deep learning. Furthermore, it enables the derivation of dimension-free Lipschitz bounds for all of the NN classes above. Another benefit of the DiPaNet architecture is that learning can now be formulated in function spaces that can be finitely parameterized using suitable bases, instead of the high-dimensional vector spaces of weights and biases.}

Compared with the neural-tangent kernel approach  \citep{Jacot2018Neural}, the developed DiPaNet
framework is purely deterministic and it preserves the nonlinear
structure of the network in the continuum limit. \ifthenelse{\boolean{short}}{}{Furthermore, it yields
$\varepsilon$-approximation guarantees directly amenable to tools
from numerical analysis and distributed parameter systems theory, without requiring any distributional assumptions on the weights.}

\ifthenelse{\boolean{short}}{}{The remainder of this paper is organized as follows. First, continuous width NNs, as continuous neural networks with an infinite number of neurons, are considered in Section~\ref{sec:2}. Then continuous depth NNs, as ordinary neural differential equations, are studied in Section~\ref{neural:ode:sec}. An unifying class of NNs, i.e., the DiPaNet with hidden distributed parameter states, is defined in Section~\ref{sec:4}. After some concluding remarks in Section~\ref{sec:5}, an appendix collects several proofs.}
\ifthenelse{\boolean{short}}{Due to space limitation, the proofs are omitted in this conference paper. See the appendix of \citep{prieur2025unified} for the proofs.}{}
\section{Continuous Width Neural Networks}
\label{sec:2}
\ifthenelse{\boolean{short}}{}{In this section we formalize the relation between discrete NNs and continuous NNs, generalizing the results of \citep{cnn} in several directions.}
\subsection{DeepNet and DeepCNN}

Let us first define a Deep neural Network (DeepNet) with $\ell$ hidden layers and $n$ neurons for each hidden layer. To do that, for $i=0,\ldots, \ell-1$ and $j=1,\ldots, n$, let $\sigma_{i+1}^{j}$ be a scalar activation function such as ReLU, hyperbolic tangent, or a Tauber-Wiener function, see, e.g.,  \citep{magon:ejc:2021,yin2021stability} for more details.
Inspired by the study on reservoir computing dynamics as introduced in \citep{grigoryeva2018universal,jaeger2007echo} for one hidden layer only, we denote the input by $X$ in $\RR^p$ and the output by $Y$ in $\RR^q$, so that the DeepNet is modeled as,
\begin{equation} \label{12:juin.2}
\begin{split}
Z_0&= L X,\\
Z_{i+1}& =  \sigma_{i+1} (W_{i+1}  Z_i) , \quad \forall i=0,\ldots, \ell-1, \\
Y&= PZ_\ell
\end{split}
\end{equation}
where $Z_0, Z_1, \ldots, Z_\ell$ are the hidden states in $\RR^n$, and $W_{1}, \ldots, W_{\ell}$, are matrices of hyperparameters in $\RR^{n\times n}$. In \eqref{12:juin.2}, $\sigma_{i+1}:\RR^n\rightarrow \RR^n$ is a componentwise activation function, that satisfies, for all $X$ in $\RR^n$ and for all $j=1,\ldots, n$,
\begin{equation}\label{eq:sig:com}
\sigma_{i+1}^{(j)} (X)=\sigma_{i+1}^{j} (X ^{(j)} ),
\end{equation}
where $\sigma_{i+1}^{(j)} (X)$ and $X ^{(j)}$ are respectively the $j$-th component of the vectors $\sigma_{i+1} (X)$ and $X$. The initial hidden state is $Z_0$ and the output is $Y$. They depend on the matrices $L\in\RR^{n\times p}$ which defines the input layer and is called the lifting matrix (or input-to-hidden matrix), and $P\in\RR^{q\times n}$ which defines the output layer and is the projection matrix (or hidden-to-output matrix). In (\ref{12:juin.2}) we assumed without loss of generality that there is no bias in any neurons (up to replacing the states $Z_i$ by $[Z_i^\top,1]^\top$).

\begin{rem}\label{rem:6:mai}
It is also possible to assume the existence of a unique activation function $\bar \sigma$ such that, for all $j$-component, \begin{equation}
\label{eq:rem:1}
\sigma_{i+1}^{(j)}(\cdot)= \bar\sigma(\cdot) \  ,
\end{equation}
in \eqref{12:juin.2}, i.e.,  this allows activation functions that do not depend on the layer index $i$. \ifthenelse{\boolean{short}}{}{This assumption simplifies some of the statements below by omitting certain parameters of the activation functions.} In the remainder of the paper, nevertheless, we consider the most general case of different activation functions.\end{rem}

Now we are in position to describe the NN with a large (or infinite) number of neurons, before considering the large depth case (with respect to the layers) in Section~\ref{neural:ode:sec}.
To do that, a second class of neural networks, that is very important for this paper, is the continuous neural network with $\ell$ hidden layers and infinitely many neurons in each hidden layer. They are modeled as:
\begin{equation}\label{7:juillet.1}
\begin{split}
\cZ_0(\tau)&= \cL(\tau) X,  \quad \forall \tau\in [0,1), \\
\cZ_{i+1}(\tau ) & =
   \csigma_{i+1} \left(\tau, \int_0^1 \cW _{i+1}(\tau,s )  \cZ_i(s)\ \mathrm{d}s\right),   \\
  &\qquad \forall \tau\in [0,1),\; \forall i=0,\ldots, \ell-1, \\
Y&=  \int_0^1\cP (\tau )\cZ _\ell (\tau) \ \mathrm{d}\tau,
\end{split}
\end{equation}
where $\cL(.):[0,1)\rightarrow \RR^{1\times p}$ and $\cP(.):[0,1)\rightarrow\RR^{q\times 1} $ are two continuous matrix functions. Moreover, in \eqref{7:juillet.1}, $\cW_{1},\cW_{2},\ldots, \cW_{l}$: $[0,1)^2\rightarrow \RR$ and $\csigma_1$, $\csigma_2$, $\ldots$, $\csigma_\ell:[0,1)\times \RR\rightarrow \RR$ are continuous functions. The hidden states $\cZ_0, \cZ_1, \ldots, \cZ_\ell$: $[0,1)\rightarrow\RR$ are functions. The basic intuition behind \eqref{7:juillet.1} is that the number of neurons is infinite and each neuron is labeled by the variable $\tau$ which lies in the set $[0,1)$, whereas each neuron in the NN \eqref{12:juin.2}
is labeled by the index of the component of $Z_i$ in $\RR^n$.
The rationale behind the integral in the second and third lines of
\eqref{7:juillet.1} is to replace the discrete sum over the neuron index, when computing the product $W_{i+1}Z_i$ in the second line of \eqref{12:juin.2}, by a continuous sum. By doing that the dimension of the hidden state $Z_i$ in \eqref{12:juin.2} is replaced by the dependence of the hidden state $\cZ_i$ with respect to a continuous variable in \eqref{7:juillet.1}.

For the case of one hidden layer only ($\ell=1$), \eqref{7:juillet.1} is particularized as
\begin{equation}\label{deep:cnn:sec:one:layer}
Y=  \int_0^1\cP (\tau )  \csigma _1\left(\tau, \int_0^1 \cW _{1}(\tau,s )  \cL(s) X\ \mathrm{d}s \right)   \ \mathrm{d}\tau.
\end{equation}
Furthermore for the case where $\csigma_1$ does not depend on the neuron, i.e. does not depend on $\tau$, and where
the function $\cW _{1}$ is defined by, for all $(\tau,s)\in [0,1)^2$,
\ifthenelse{\boolean{short}}{\begin{equation}\label{deep:cnn:sec:one:layer:hyper}
\cW _{1}(\tau,s)
=0\mbox{ if }\tau\neq s\ ,
\cW _{1}(\tau,\tau)
=1\ ,
\end{equation}}{\begin{equation}\label{deep:cnn:sec:one:layer:hyper}
\begin{split}
\cW _{1}(\tau,s)&=0\mbox{ if }\tau\neq s\ ,
\\
\cW _{1}(\tau,\tau)&=1\ ,
\end{split}
\end{equation}}
the NN (\ref{deep:cnn:sec:one:layer}) becomes
\begin{equation}
\label{deep:cnn:sec:one:layer:cnn}
Y=  \int_0^1\cP (\tau )  \csigma_1 (  \cL(\tau) X)  \ \mathrm{d}\tau,
\end{equation}
which is exactly the Continuous Neural Network (CNN) defined by  \citep{cnn} (see Eq. (2) therein, with integration over a bounded set $E$).
Therefore, the NN in \eqref{7:juillet.1} generalizes CNNs for the case of several hidden layers, and thus\ifthenelse{\boolean{short}}{}{, in analogy with Deep Neural Networks considered in \citep{yamasaki1993lower,yun2019small},} we may call neural networks described by \eqref{7:juillet.1} DeepCNNs.
\begin{rem}
The CNN of \citep{cnn} is recovered from the DeepCNN~\eqref{7:juillet.1} as the
special case $\ell = 1$ with $W_1$ satisfying~\eqref{deep:cnn:sec:one:layer:hyper} and an  activation function independent of $\tau$.
The DeepCNN generalizes this in three directions:
\emph{(i)}~ $W_i(\tau, s)$ is a general continuous function on $[0,1)^2$, allowing neuron $\tau$ to receive input from all neurons
$s$ within the same layer; \emph{(ii)}~the activation function $\sigma_i(\tau, \cdot)$ may depend on
both the layer index $i$ and the neuron index $\tau$;
\emph{(iii)}~$\ell \geq 1$ hidden layers are considered.
\end{rem}
\begin{rem}
In \eqref{7:juillet.1}, the hidden states lie in a the set of continuous functions from $[0,1)$ to $\RR$, which is an infinite-dimensional space, whereas the input and the output are in finite-dimensional vector spaces, respectively $\RR^p$ and $\RR^q$. We could also consider infinite-dimensional input and output spaces by removing the first and the last lines in \eqref{7:juillet.1}, and seeing $Z_0$ as a input and $Z_\ell$ as the output. However to ease the presentation and link the DeepCNN modeled by \eqref{7:juillet.1} with the DeepNet modeled by \eqref{12:juin.2}, we prefer to consider only finite-dimensional input and output vector spaces in this paper.
\end{rem}
\ifthenelse{\boolean{short}}{}{The works \citep{cnn} and  \citep{ruthotto2020deep} link continuous NNs and standard finite-dimensional NNs in the case of one hidden layer only, by considering the limit as the number of neurons goes to infinity.} Next, we will derive a result that links the DeepCNN with finite-dimensional Deep neural Networks as described by the DeepNet. To do that, we need to introduce the notion of a sequence of neurons with hyperparameter matrices $W_{i}$ and activation functions that are ``increasing'' as done in the following definition. This definition is not needed in \citep{cnn} because the same activation function is used for all neurons and because only hyperparameter matrices as in \eqref{deep:cnn:sec:one:layer:hyper} are considered in this reference. \ifthenelse{\boolean{short}}{}{This definition is related to tensors with increasing dimension as done in \citep{matsui2024broadcastproductshapealignedelementwise} dealing with broadcasting product between tensors.}
\begin{defn}
    Let $\ell$ be a positive integer.
    \newline
    $\bullet$ Let $(L^n)_{n>0}$, for all $i=1,\ldots, \ell$, $(W^n_i)_{n>0}$ and $(P^n)_{n>0}$ be sequences of matrices such that, for all $n\in\NN\setminus \{0\}$, $L^n$ is in $\RR^{n\times p}$, $W_i^n$ is in $\RR^{n\times n}$ and $P^n$ is in $\RR^{q\times n}$. We say that these sequences of matrices are {\em consistent} if given two integers $0<n\leq n'$, it holds
\begin{equation*}
\begin{split}
&L^n= \Pi_{n\leftarrow n' }L^{n'}\ ,\quad  P^n\Pi_{n\leftarrow n' }= P^{n'}\ ,
\\
&W^n_i\Pi_{n\leftarrow n' }= \Pi_{n\leftarrow n' }W^{n'}_i\ , \quad \forall  i=1,\ldots, \ell\ ,
\end{split}
\end{equation*}
where $\Pi_{n\leftarrow n' }$ is in $\RR^{n\times n'}$ and is the projection matrix from $\RR^{n'}$ to $\RR^n$:
$$
\Pi_{n\leftarrow n' }= \begin{bmatrix}I_{n} & 0\end{bmatrix}\ ,
$$
with $I_n$ stands for the identity matrix of size $n$.
\newline
$\bullet$ Let, for all $i=1,\ldots, \ell$, $(\sigma ^n_i)_{n>0}$ be a sequence of functions: $\RR^n\rightarrow \RR^n$. We say that this sequence of activation functions is {\em consistent} if given two integers $0<n\leq n'$, it holds, for all $i=1,\ldots, \ell$,
$$
\sigma^n_i(\Pi_{n\leftarrow n' }Z)= \Pi_{n\leftarrow n' }\sigma^{n'}_i (Z)\ , \quad \forall Z\in \RR^{n'}\ ;
$$
$\bullet$
The sequences are of {\em bounded variation} if, for all $i=1,\ldots, \ell$, for all $1\leq m\leq n$,
$$\begin{array}{c}
\sum_{n >0}\|L^{n+1(n+1:)}-L^{n(n:)}\| <\infty\ ,
\\
\sum_{n >0}|W_i^{n+1(n+1,m)}-W_i^{n(n,m)}|\\
\qquad \qquad+|W_i^{n(n,m)}-W_i^{n(n,m-1)}| <\infty\ ,
\\
\sum_{n >0}\sup_{\RR^{n+1}}\|\Pi_{n\leftarrow n+1 }\sigma_i^{n+1}(.)-
\sigma_i^{n}(\Pi_{n\leftarrow n+1 }.)\| <\infty\ ,\\
\sum_{n >0}\|P^{n+1(:n+1)}-P^{n(:n)}\| <\infty\ ,
\end{array}
$$
where $L^{n(n:)}$ denotes the $n$-row of the matrix $L^{n}$, $W_i^{n(a,b)}$ denotes the $(a,b)$-term in the matrix $W_i^n$, and $P^{n(:n)}$ denotes the $n$-column of the matrix $P^n$.
\end{defn}

Given a sequence $(l^n)_{n>0}$ in  $\RR^{1\times p}$, a consistent sequence of matrices $(L^n)_{n>0}$ is given by
\begin{equation}\label{6:mai}
L^n= \begin{bmatrix}
    l^1\\ l^2 \\ \vdots \\ l^n
    \end{bmatrix}
\in \RR^{n\times p}\ .
\end{equation}
Reciprocally, given a consistent sequence of matrices $(L^n)_{n>0}$ in $\RR^{n\times p}$, defining $l^n$ in $\RR^{1\times p}$ by $l^1=L^1$ and, for all $n\geq 2$, $l^n$ by
$$
L^n :=\begin{bmatrix}
    L^{n-1} \\ l^n
    \end{bmatrix}\ ,
$$
we can check that (\ref{6:mai}) is satisfied. Moreover, assuming, as in Remark \ref{rem:6:mai}, that there exists a unique activation function $\bar \sigma$ such that (\ref{eq:rem:1}) holds, then the sequence of activation functions $(\sigma^n)_{(n>0)}$ is consistent.

\ifthenelse{\boolean{short}}{}{The last part of this definition deals with the increments of consecutive terms. Intuitively, a sequence of hyperparameters is of bounded variation if the sum of all increments is bounded. This is needed to define the homogenization part of our first main result.}

We are now in a position to state the links between the DeepCNN and the DeepNet as follows:
\begin{thm}\label{th:1}
Let $\ell$ be a positive integer.
\newline
$\bullet$ Given a DeepCNN (\ref{7:juillet.1})
and a positive integer $n$, its {\em discretization}  into $n$ neurons is a DeepNet modeled by \eqref{12:juin.2}.
More specifically, given uniformly continuous functions $\cL$, $\cP$, $\cW_{i}$, continuous activation functions $\csigma_{i}$, $i=1,\ldots, \ell$, and positive values $r$ and $\varepsilon$,
there exists a sufficiently large integer $\overline{n}$ satisfying, for all $n\geq\overline{n}$, the existence of matrices $L$, $P$, and $W_{i}$, and activation functions $\sigma_i$, $i=1,\ldots, \ell$, so that for all $X\in \RR^p$ satisfying $\|X\|\leq r$, the distance between the output $Y_{\eqref{12:juin.2}}\in \RR^q$ of \eqref{12:juin.2} with $n\geq \overline{n}$ neurons and the output $Y_{(\ref{7:juillet.1})}\in \RR^q$ of (\ref{7:juillet.1}) is less than $\varepsilon$:
\begin{equation}
\label{eq:discreti}
\forall n\geq \overline{n}, \|X\|\leq r \Rightarrow \| Y_{(\ref{7:juillet.1})}- Y_{\eqref{12:juin.2}}\|\leq \varepsilon\ .
\end{equation}
\newline
$\bullet$ Conversely the {\em homogenization} of a DeepNet as given by (\ref{12:juin.2}), as $n$ tends to infinity, is a DeepCNN modeled by (\ref{7:juillet.1}).
More specifically, assuming consistent sequences of matrices $(L^n)_{n>0}$, $(P^n)_{n>0}$, $(W^n_i)_{n>0}$ with bounded variation, a consistent set of activation functions $(\sigma^n_i)_{n>0}$ with bounded variation, for $i=1,\ldots, \ell$, and positive values $r$ and $\varepsilon$, there exist functions $\cL$, $\cP$, $\cW_i$, and $\csigma_i$, for $i=1,\ldots, {\ell}$, and a sufficiently large integer $\overline{n}$ satisfying the property, for all $X\in\RR^p$ satisfying $\|X\|\leq r$, the distance between the outputs of \eqref{12:juin.2} with $n\geq \overline{n}$ neurons and of (\ref{7:juillet.1}) is less than $\varepsilon$:
$$
\forall n\geq \overline{n}, \|X\|\leq r \Rightarrow \| Y_{(\ref{7:juillet.1})}- Y_{\eqref{12:juin.2}}\|\leq \varepsilon
\ .$$
\end{thm}

Intuitively, in Theorem \ref{th:1}, the positive value $r$ is the radius of admissible inputs for the NN under consideration, and $\varepsilon$ is the corresponding bound of the error between NNs. In this result $r$ could be as large as we want (but finite) and $\varepsilon$ as small as needed (but positive). Moreover the number of hidden layers is fixed and given by $\ell$. \ifthenelse{\boolean{short}}{}{The proof of this theorem is given in Appendix, Section \ref{sec:pro:th:1}.}
\begin{rem}
Theorem~\ref{th:1} provides a rigorous counterpart to the informal  discretization argument of \citep{cnn}, extending it to the deep case $\ell \geq 1$ via a bidirectional result
covering both discretization and homogenization, with
$\varepsilon$-approximation guarantees that track the
error propagation through all $\ell$ hidden layers.
\end{rem}

\subsection{DeepResNet and DeepResCNN}
\label{sec:deepresnet}

The Deep Residual neural Network (DeepResNet) is a slight modification of DeepNet \eqref{12:juin.2}, by modifying the recurrent equation in (\ref{12:juin.2})  as follows (see \citep{he2016deep}):
\begin{equation} \label{12:juin.3}
\begin{split}
Z_0&= L X\ ,\\
Z_{i+1}& =  Z_i+\sigma_{i+1} (W_{i+1}  Z_i) , \quad \forall i=0,\ldots, \ell-1\ , \\
Y&= PZ_\ell\ ,
\end{split}
\end{equation}
using similar notation as in (\ref{12:juin.2}). Another variant of this model is obtained by adding a direct link from the first hidden layer to the output map as follows:
\begin{equation}\label{5:dec}
\begin{split}
Z_0&= L X\ ,\\
Z_{i+1}& =  Z_i+\sigma_{i+1} (W_{i+1}  Z_i) , \quad \forall i=0,\ldots, \ell-1\ , \\
Y&= P[Z_\ell^\top,Z_0^\top]^\top\ ,
\end{split}
\end{equation}
where $P$ is a matrix in $\RR^{q\times 2n}$. To be more specific, \citep{he2016deep} introduces a direct link from the first input to the output, and replaces the third line of  \eqref{5:dec} by
$Y= P[Z_\ell^\top,X]^\top$. Note that \citep{he2016deep} does not have any lifting matrix $L$, so this is consistent with \eqref{5:dec}.

Inspired by \eqref{5:dec}, we can adapt the definition of Deep Continuous Neural Networks to get the notion of Deep  Residual Continuous Neural Networks (DeepResCNN):
\begin{equation} \label{12:juin.4}
\begin{split}
\cZ_0(\tau)&= \cL(\tau) X,  \quad \forall \tau\in [0,1)\ ,\\
\cZ_{i+1}(\tau)& =  \cZ_i(\tau)+\csigma _{i+1}\left(\tau,\int_0^1 \cW_{i+1} (\tau,s)  \cZ_i(s)\ \mathrm{d}s\right)   \ , \\
&\qquad \forall \tau\in [0,1), \; \forall i=0,\ldots, \ell-1 \ , \\
Y&=  \int_0^1 \cP (\tau )[\cZ _\ell (\tau)^\top \cZ _0 (\tau)]^\top   \ \mathrm{d}\tau \ .
\end{split}
\end{equation}
We can derive a similar result to Theorem \ref{th:1} by comparing the DeepResNet and the DeepResCNN, as follows.
\begin{cor}\label{cor:1res}
Let $n$ and $\ell$ be two positive integers.
\newline $\bullet$ The {\em discretization} of a DeepCNN as given by  (\ref{12:juin.4}) is given as DeepResNet as given by \eqref{5:dec}.
\newline $\bullet$ The {\em homogenization} of any sequence of DeepResNet as given by \eqref{5:dec} (with consistent sequences of matrices) is a DeepCNN when the number of neurons increase and go to infinity.
\end{cor}

\section{Continuous Depth Neural Networks}
\label{neural:ode:sec}

\subsection{DeepResNet and NeuralODE}

In addition to the classes of NNs introduced in the previous section, another class of neural networks that is useful in this paper is the Neural Ordinary Differential Equations (NeuralODE) that is also a limit case but when the number of hidden layers tends to infinity, as introduced in \citep{neuralODE}. They are modeled by
\begin{equation}\label{7:juillet.3.2}
\begin{split}
\cZ(0)&= L X,\\
\dot \cZ(t)& =
 \csigma (t,\cW(t) \cZ(t)), \quad\forall t\in[0,T]
 \ ,
 \\ Y &
 = P \cZ(T)\ ,
\end{split}
\end{equation}
where $T>0$ is given, $\cW: [0,T]\rightarrow \RR^{n\times n}$ is a continuous matrix function, $L\in\RR^{n\times p}$ and $P\in\RR^{q\times n}$ are two matrices, and $ \csigma :[0,T]\times\RR^n\rightarrow\RR^n$ is a continuous function such that, for all  $t\in [0,T]$,  $\csigma(t,.)$ is a componentwise activation function as in \eqref{eq:sig:com}. An alternative definition is given by
\begin{equation}\label{7:juillet.3.2.2}
\begin{split}
\cZ(0)&= L X,\\
    \cZ(t) &= \cZ(0) + \int_0^t  \csigma (s,\cW (s) \cZ(s)) \ \mathrm{d}s  , \quad\forall t\in[0,T] \  , \\
     Y&= P\cZ(T)\ .
\end{split}
\end{equation}

Both definitions of NeuralODE are equivalent as stated in the following result.
\begin{lem}\label{lem:0}
The expressions (\ref{7:juillet.3.2}) and (\ref{7:juillet.3.2.2}) are equivalent.
\end{lem}
This lemma directly follows from the expression of the solution to the Cauchy problem
\begin{equation}
\nonumber
\begin{split}
\cZ(0)&= L X,\\
\dot \cZ(t)& =
 \csigma (t,\cW(t) \cZ(t)), \quad\forall t\in[0,T]
 \ .
\end{split}
\end{equation}

The links from
 DeepResNet and NeuralODE are sketched in \citep{neuralODE} (see also \citep{LuZhong2018BeyondFinite} and \citep{Weinan2017ProposalMLviaDS}). The rationale behind the link from NeuralODE to  DeepResNet is the use of a numerical scheme as the Euler discretization method. To study the converse, that is to start from a sequence and to see it as an Euler discretization, asks introducing the following definition.
 \begin{defn}
Let $T>0$. Let $(W_1^\ell, W_2^\ell,\ldots, W_\ell^\ell)_{\ell>0}$ be a sequence so that for all $\ell>0$ and for $i=1,\ldots, \ell$, $W_i^\ell$ is a matrix in $\RR^{n\times n}$. Then we say that the sequence $((W_i^\ell)_{i =1,\ldots, \ell})_{l>0}$ is
\newline
$\bullet$
{\em uniformly bounded} if
$$
\sup_\ell\sup_{i=1,\ldots, \ell}\| W_i^\ell\| <\infty\ ; $$
\newline
$\bullet$
{\em equicontinuous} if the sequence of functions $(\cW_\ell)_{\ell>0}$ is equicontinuous\footnote{Let us recall that the sequence of functions $(\cW_\ell)_{\ell>0}$ is said to be equicontinuous on $[0,T]$ if, for all $t$ in $[0,T]$ and for all $\epsilon >0$, there exists $\delta>0$ such that, for all $s$ in $[0,T]$ satisfying $|t-s|\leq \delta$ and for all $\ell>0$, it holds $\| \cW_\ell(s)-\cW_\ell(t)\|\leq \epsilon$.} on $[0,T]$,
where, for each $\ell$, the matrix function $\cW_\ell:[0,T]\rightarrow \RR^{n\times n}$ is the
interpolant function, that is defined by, for all $i=0,\ldots, \ell-1$,
$$
\cW_\ell(\frac{i}{\ell})=
W_i^\ell\ ,
$$
and that is linear on each interval $
[\frac{i}{\ell}T,\frac{i+1}{\ell}T]$.

Let $(\sigma_1^\ell, \sigma_2^\ell,\ldots, \sigma_\ell^\ell)_{l>0}$ be a sequence so that for all $\ell>0$ and $i=1,\ldots, \ell$, $\sigma_i^\ell : \RR^{n}\rightarrow \RR^{n}$ is an activation function. Then we say that the sequence $((\sigma_i^\ell)_{i =0,\ldots, \ell-1})_{l>0}$ is
\newline
$\bullet$
{\em uniformly bounded} if
$$
\sup_\ell\sup_{i=1,\ldots, \ell}\sup_{Z\in\RR^n} \| \sigma _i^\ell(Z)\| <\infty$$
\newline
$\bullet$
{\em equicontinuous} if the sequence of functions $(\csigma_\ell)_\ell$ is equicontinuous on $[0,T]$,
where, for each $\ell$, the function $\csigma_\ell:[0,T]\times \RR^n\rightarrow \RR^{n}$ is the
interpolant function, that is defined by, for all $i=0,\ldots, \ell-1$ and for all $Z\in\RR^n$,
$$
\csigma_\ell(\frac{i}{\ell},Z)=
\sigma_i^\ell(Z)\ ,
$$
and that is linear with respect to its first argument on each interval $
[\frac{i}{\ell}T,\frac{i+1}{\ell}T]$.
\end{defn}

\begin{thm}\label{th:2}
Let $n$ be a positive integer.
\newline
$\bullet$ Given a NeuralODE (\ref{7:juillet.3.2.2}) and a positive integer $\ell$, its {\em discretization} into $\ell$ hidden layers is a DeepResNet modeled by \eqref{12:juin.3}.
More specifically, given matrices $L\in\RR^{n\times p } $, $P\in\RR^{q\times n }$, continuous functions $\cW$ and $\csigma$, and positive values $r$ and $\varepsilon$, there exists a sufficiently large integer $\overline{\ell}$ satisfying, for all $\ell\geq\overline{\ell}$ the existence of  matrices $W_{i}$, and activation functions $\sigma_i$, $i=1,\ldots, \ell$, so that for all $X\in \RR^p$ with $\|X\|\leq r$, the distance between the output $Y_{\eqref{12:juin.3}}\in \RR^q$ of \eqref{12:juin.3} with $\ell\geq \overline{\ell}$ hidden layers and the output $Y_{(\ref{7:juillet.3.2.2})}\in \RR^q$ of (\ref{7:juillet.3.2.2}) is less than $\varepsilon$:
\begin{equation}
\label{eq:discreti:2}
\forall \ell\geq \overline{\ell}, \|X\|\leq r \Rightarrow \| Y_{(\ref{7:juillet.3.2.2})}- Y_{\eqref{12:juin.3}}\|\leq \varepsilon\ .
\end{equation}

$\bullet$ Conversely the {\em homogenization} of a DeepResNet as given by (\ref{12:juin.3}), as $\ell$ tends to infinity, is a NeuralODE modeled by (\ref{7:juillet.3.2.2}).
More specifically, given matrices $L\in\RR^{n\times p } $, $P\in\RR^{q\times n }$, an uniformly bounded and equicontinuous sequence of matrices $((W^\ell_i)_{i=1,\ldots,\ell})_{\ell>0}$, an uniformly bounded and equicontinuous sequence of functions $((\ell\sigma^\ell_i)_{i=1,\ldots,\ell})_{\ell>0}$ and positive values $r$ and $\varepsilon$.

Then there exist functions $\cW$ and $\csigma$ and a sufficiently large integer $\overline{\ell}$ satisfying the property, for all $X\in\RR^p$ satisfying $\|X\|\leq r$ the distance between the outputs of \eqref{12:juin.3} with $\ell\geq \overline{\ell}$ hidden layers and of (\ref{7:juillet.3.2.2}) is less than $\varepsilon$:
$$
\forall \ell\geq \overline{\ell}, \|X\|\leq r \Rightarrow \| Y_{(\ref{7:juillet.3.2.2})}- Y_{\eqref{12:juin.3}}\|\leq \varepsilon
\ .$$
\end{thm}
\ifthenelse{\boolean{short}}{}{Intuitively, in Theorem \ref{th:2}, the number of neurons is fixed and given by $n$.}\ifthenelse{\boolean{short}}{}{The proof of this theorem is given in Appendix, Section \ref{sec:pro:th:2}.}
\subsection{DeepResNet and NeuralResODE}

Following the computations done in Section \ref{sec:deepresnet}, we may adapt the definitions of NeuralODE as written in (\ref{7:juillet.3.2}) and (\ref{7:juillet.3.2.2}) by considering the residual case. To do that, we just need to replace the last line of (\ref{7:juillet.3.2}) (respectively (\ref{7:juillet.3.2.2})) by
$$
Y= P[\cZ(T)^\top,\cZ(0)^\top]^\top\ .
$$
Thus we may define the class of Neural Residual Ordinary Differential Equations (NeuralResODE) and we may prove
a similar result as in Theorem~\ref{th:2} for this class of NNs, by comparing NeuralResODEs with the limit of DeepResNets as the number of hidden layers increases to infinity.

\section{The DiPaNet Representation}
\label{sec:4}

\subsection{DeepResNet and DiPaNet}

In this section we describe a new input/output NN architecture that unifies all previous NNs and their approximation when both the number of neurons and of hidden layers increase to infinity. More precisely the new class of NNs is given by, for all
$\tau\in [0,1)$ and for all $t\in[0,T]$, \begin{equation}\label{12:juin}
\begin{split}
\cZ(\tau,0)&= \cL(\tau) X \ ,\\
   \cZ(\tau,t)& = \cZ(\tau,0)
   + \int _0^t  \csigma (\tau,s, \int_0^1 \cW (\tau , u,s)  \cZ(u,s)\mathrm{d}u)  \mathrm{d}s  ,
\\
Y&=  \int_0^1 \cP (\tau )Z (\tau,T)\  \mathrm{d}\tau \ ,
\end{split}
\end{equation}
where $T>0$ is given, $X$ in $\RR^p$ is an input vector, $\cZ:[0,1)\times [0,T]\rightarrow \RR $ is the hidden state, and the output vector is
$Y$ in $\RR^q$. In (\ref{12:juin}),
$\cL(.):[0,1)\rightarrow \RR^{1\times p}$ is the lift matrix function and $\cP(.):[0,1)\rightarrow \RR^{q\times 1}$ is the projection matrix function. This model depends on a hyperparameter function $\cW:[0,1)\times[0,1)\times [0,T]\rightarrow \RR$. Finally the function
$ \csigma :[0,1)\times [0,T]\times\RR\rightarrow\RR$ is a function such that, for all  $(\tau,t)\in [0,1)\times [0,T]$,  $\csigma(\tau,t,.)$ is a scalar activation function. Note that, as soon as the functions $\cL$, $\csigma$, $\cW$ and $\cP$ are continuous, given an input vector $X$ in $\RR^p$, there exists a unique output vector $Y$ in $\RR^q$ satisfying \eqref{12:juin}.

An alternative representation of the previous model is given by the following integro-differential equation, for all $\tau\in [0,1)$ and for all $t\in [0,T]$,
\begin{equation}\label{12:juin.t}\begin{split}
\cZ(\tau,0)&= \cL(\tau) X\ ,\\
\partial_t \cZ(\tau, t) &=  \csigma \left(\tau,t,\int_0^1\cW (\tau , s,t)  \cZ(s ,t)\ \mathrm{d}s\right)  \ , \quad\forall t\in[0,T]\ ,
\\
Y&=  \int_0^1 \cP (\tau )\cZ (\tau,T)  \mathrm{d}\tau\ .
\end{split}
\end{equation}

\begin{lem}
The expressions (\ref{12:juin}) and (\ref{12:juin.t}) are equivalent.
\end{lem}
\ifthenelse{\boolean{short}}{}{The proof of the previous lemma follows from the proof of Lemma \ref{lem:0} and is thus skipped.
\begin{rem}
A neural integro-differential equation similar to (\ref{12:juin.t}) was also derived in \citep{Zappala2023NeuralIntegroDifferential}, albeit with a different purpose, i.e., modeling of non-local dynamics represented by integro-differential equations, and with a different construction of an integral kernel operator. Furthermore, \citep{Zappala2023NeuralIntegroDifferential} did not establish formal relations with classical NNs such as DeepResNets and other NNs as considered in Sections \ref{sec:2}-\ref{neural:ode:sec} of this paper.
\end{rem}}
The above-defined class of NNs is a distributed parameter system because the state $Z$ depends on the time, and of functions depending themselves on both variables $t$ and $\tau$. We call such a NN as a Distributed Parameter neural Network (DiPaNet). \ifthenelse{\boolean{short}}{}{The links of this NN with respect to the other ones previously studied in this paper lie in the
discretization (or conversely in homogenization) in width. This is performed via standard numerical discretization (or conversely by interpolation) in the domain $[0,1)$ to approximate the integral from $0$ to $1$ in \eqref{12:juin.t}, as done in the proof of Theorem \ref{th:1} when dealing with DeepCNN. Another link with one previously considered NN is the discretization (or conversely homogenization) in depth that is performed via the dynamical system perspective in  \eqref{12:juin.t}, as done in the proof of Theorem \ref{th:2} dealing with NeuralODE, rather than numerical discretization of the integral from $0$ to $T$ in \eqref{12:juin} for example.}

The connection between the neural networks DeepResCNN (\ref{12:juin.4}) and DiPaNet (\ref{12:juin.t}) is made precise\ifthenelse{\boolean{short}}{:}{in the following result (whose proof is skipped because it is similar to the proof of Theorem \ref{th:2}):}
\begin{thm}\label{lem:1} Let $\ell$ be a positive integer.
\newline
$\bullet$
The {\em discretization} of a DiPaNet as given by (\ref{12:juin.t}) into $\ell$ hidden layers is a
DeepResCNN modeled by \eqref{12:juin.4}.
\newline
$\bullet$
The {\em homogenization} of a DeepResCNN as given by (\ref{12:juin.4}), as $\ell$ tends to infinity, is a DiPaNet modeled by (\ref{12:juin.t}).
\end{thm}

The link between the neural networks NeuralODE (\ref{7:juillet.3.2}) and DiPaNet (\ref{12:juin.t}) is made precise in the following result\ifthenelse{\boolean{short}}{:}{(whose proof is skipped because it is similar to the proof of Theorem \ref{th:1}):}
\begin{thm}Let $n$ be a positive integer.
\newline
$\bullet$
The {\em discretization} of a DiPaNet as given by (\ref{12:juin.t}) into $n$ neurons is a
NeuralODE modeled by (\ref{7:juillet.3.2}).
\newline
$\bullet$
The {\em homogenization} of a NeuralODE as given by (\ref{7:juillet.3.2}), as $n$ tends to infinity, is a DiPaNet modeled by (\ref{12:juin.t}).
\label{lem:2}
\end{thm}

Since the discretization of a DiPaNet into $n$ neurons is a NeuralODE (by Theorem \ref{lem:2}), and since the discretization of a NeuralODE into $\ell$ hidden layers is a DeepResNet (by Theorem \ref{th:2}), we get the following result:
\begin{cor}\label{cor:1}
Let $n$ and $\ell$ be two positive integers.
\newline
$\bullet$ The {\em discretization} of a DiPaNet as \ifthenelse{\boolean{short}}{in}{given by} (\ref{12:juin.t}) into $n$ neurons and $\ell$ hidden layers is a
DeepResNet modeled by (\ref{12:juin.3}).
\newline
$\bullet$ The {\em homogenization} of a DeepResNet as given by (\ref{12:juin.3}), as $\ell$ and $n$ tend to infinity, is a DiPaNet modeled by (\ref{12:juin.t}).
\end{cor}
\ifthenelse{\boolean{short}}{}{One way to prove this corollary is to use Theorems \ref{lem:2} and \ref{th:2}. An alternative approach is to use Theorems \ref{lem:1} and \ref{th:1}.}

The links between the neural networks architectures studied in this paper are sketched in Figure \ref{table:1}. Additionally, DeepCNNs and DeepNets are linked similarly as DeepResCNNs and DeepResNets, as stated in Theorem~\ref{th:1}.

\begin{figure}
\begin{center}\begin{tabular}{|ccc|}
\hline
DeepResCNN \eqref{12:juin.4} &
$\begin{array}{c}{\color{blue}\longleftarrow}\\
{\color{red}\longrightarrow}\end{array} $& {\bf DiPaNet (\ref{12:juin.t})}
\\
${\color{blue}\big\downarrow}\;{\color{red}
\big\uparrow}$  &&${\color{blue}\big\downarrow}\;{\color{red}
\big\uparrow}$
\\
DeepResNet (\ref{12:juin.3})&$\begin{array}{c}{\color{blue}\longleftarrow}\\
{\color{red}\longrightarrow}\end{array} $ &NeuralODE (\ref{7:juillet.3.2})
\\\hline
\end{tabular}
  \caption{Sketch of {\color{blue}discretizations} (${\color{blue}\big\downarrow}$ and ${\color{blue}\longleftarrow}$ arrows) and {\color{red}homogenizations} (${\color{red}
\big\uparrow}$ and ${\color{red}\longrightarrow}$
arrows).}
\label{table:1}
\end{center}
\end{figure}

\subsection{DiPaResNet and further generalizations}

The class of NNs given by the DiPaNet as modeled by (\ref{12:juin})-\eqref{12:juin.t} could be generalized in several directions. First the residual case could be considered by adding direct transfer from the input $X$ to the output $Y$. It yields the following Distributed Parameter Residual neural Network (DiPaResNet), for all
$\tau\in [0,1)$ and for all $t\in[0,T]$,
\begin{equation}\nonumber
\begin{split}
\cZ(\tau,0)&= \cL(\tau) X \ ,\\
   \cZ(\tau,t)& = \cZ(\tau,0)
   + \int _0^t   \csigma (\tau,u, \int_0^1\cW (\tau , u,s)  \cZ(u,s)\mathrm{d}u) \mathrm{d}s ,
\\
Y&=  \int_0^1 \cP (\tau )[\cZ (\tau,T)^\top \cZ (\tau,0)]^\top   \ \mathrm{d}\tau \ .
\end{split}
\end{equation}

Another generalization is by considering distributed parameters spaces for the input and output sets, i.e., instead of $X\in\RR^p$ and $Y\in \RR^q$, by considering functions $X$ and $Y$, and by constructing Distributed Parameter neural Operator Networks. This will be considered in future work.
\ifthenelse{\boolean{short}}{}{\begin{rem}
The DiPaNet formulation~(\ref{12:juin})-\eqref{12:juin.t}  is strictly feedforward
in the temporal domain, and thus does not directly capture recurrent neural network (RNN) architectures. However, extending DiPaNet to the recurrent setting is conceptually possible by replacing the autonomous dynamics in~(16) with a controlled integro-differential equation driven by a
continuous input path, in the spirit of Neural Controlled Differential Equations (Neural CDEs) \citep{Kidger2020}, which have been established as the continuous-time
analogue of RNNs.
\end{rem}
\begin{rem}
The approximation results developed in this paper establish pointwise convergence of the input-output maps for fixed network parameters, i.e., they show that the forward dynamics of a DeepResNet converge to those of a DiPaNet as $n, \ell \to \infty$. This is distinct from, and weaker than, the variational question
of whether optimal trained parameters of a DeepResNet converge to optimal parameters of the limiting DiPaNet, i.e., whether
minimizers of the discrete training objective converge to minimizers of the continuous training objective. The latter
has been recently addressed for the depth direction (DeepResNet to NeuralODE) in \citep{Thorpe2023} using variational convergence techniques (i.e., $\Gamma$-convergence). Extending such results to the width direction and to the full DiPaNet framework is an important open problem left for future work.
\end{rem}}

\section{Conclusion}
\label{sec:5}
This work considered
different classes of NNs such as Deep Neural Networks, Continuous Neural Networks, Neural Differential Equations and their Residual forms. Moreover a new class of NNs has been introduced, DiPaNet, that includes a distributed parameter hidden state, and that can be characterized as the limit case of classical deep residual NNs as the number of hidden layers and/or neurons goes to infinity. In addition, the approximation of DiPaNets has been studied, yielding a deep residual NN with a finite number of hidden layers or of neurons,
and corresponding approximations of the input/output map of a DiPaNet.

\ifthenelse{\boolean{short}}{}{The developed DiPaNet architecture opens the door to many interesting research lines. The authors plan to study the properties of the input/output maps by considering the infinite-dimensional space of the hidden state. The derivation of Lipschitz constants and of other types of bounds for this input/output map will be done. Moreover, loss functions will be considered, together with the probabilistic generalization that could be useful for a learning procedure.
 Specific parameterizations of the hyperparameter functions will be considered, in particular, for control design problems for some partial differential equations. Extensions to operator learning will also be considered.}
\section*{Acknowledgments}
Mircea Lazar gratefully acknowledges the financial support received from the HORIZON-MSCA EU Project COVER, grant agreement 101086228. The authors would like to thank Epiphane Loko for useful comments on a previous version of this paper.

\ifthenelse{\boolean{short}}{}{\appendix
\section{Technical proofs}
Let us first state some lemmas.

\subsection{Proof of intermediate technical results}\label{appen:sec}

In this section we state and prove some lemmas needed for the proof of Theorem~\ref{th:1}. Let us first prove the following technical lemma.
\begin{lem}\label{lemma1}
For any $\nu>0$ and for any uniformly continuous function $\cF:[0,1)\rightarrow \RR^u$,
there exist a positive integer $n$, a positive value $\Delta \tau>0$ and a matrix $F$ in $\RR^{n\times u}$ such, for all $j=1,\ldots, n$, for all $\tau \in [j\Delta \tau,(j+1)\Delta \tau )\cap [0,1)$,
\begin{equation}\label{eq:approx:}
\|F^{(j:)}-\cF(\tau)\|\leq \nu\ ,
\end{equation}
where $F^{(j:)}$ is the $j$-th row of the matrix $F$.
\end{lem}
\begin{proofof}{ \em Lemma \ref{lemma1}.\ }
Since
$\cF$ is uniformly continuous, there exists $\Delta \tau$ such that for all $\tau, \tau^\prime$ satisfying $|\tau -\tau^\prime|\leq \Delta \tau $, it holds
\begin{equation}
\label{eq:equi:con}
\|\cF(\tau)-\cF(\tau^\prime)\|\leq \nu
\ .
\end{equation}
Let ${n}= \lfloor  \frac{1}{\Delta \tau}\rfloor+1$, and let $F\in \RR^{n+1}$ be defined by, for all $j=1,\ldots, n$,
$$
F^{(j:)}=\cF(j\Delta \tau)\ .
$$
Then \eqref{eq:approx:} follows from \eqref{eq:equi:con} and this definition of the matrix $F$ in $\RR^{n\times u}$.
\end{proofof}

Before stating the next technical lemma, let us recall that, given a bounded interval $[a,b]$ and a function $\cF : [a,b] \to \mathbb{R}^u $, the total variation of $\cF$ on $[a,b]$ is defined as:
\[
V(\cF,[a,b]) = \sup \left\{ \sum_{i=1}^n \|\cF (x_i) - \cF(x_{i-1})\| \right\},
\]
where the supremum is taken over all possible partitions \( a = x_0 < x_1 < x_n=b\) of the interval \([a,b]\).
Moreover the function \( \cF\) is said to be of {bounded variation} on $[a,b]$
if its total variation is finite $V(\cF,[a,b]) < \infty$.
Finally, if \( \cF \) is defined on \([0, \infty)\), then \( \cF \) is of bounded variation on \([0, \infty)\) if:
\[
V(\cF, [0, \infty)) := \sup_{T > 0} V(\cF, [0, T]) < \infty.
\]

\begin{lem}
\label{lemma:2}
Assume given $\epsilon>0$ and a piecewise constant function $\cF:[0,\infty)\rightarrow \RR^u$, that is, a function $\cF$ such that there exists a unbounded sequence $(x_n)_{n\in\NN}$ satisfying $x_0=0$ and $x_n<x_{n+1}$ for all $n\in\NN$, and a sequence $(c_n)_{n\in\NN}$ in $\RR^u$ such that, for all \( x \in [x_n, x_{n+1}) \),
\( \cF(x) = c_n\ . \)
 Moreover, assume that the function $\cF$ is of bounded variation.
\newline
Then there exists a continuous function $\tilde \cF:[0,\infty)\rightarrow \RR^u$ such that
$
\int_0^\infty
\|\cF(s)-\tilde \cF(s)\|\ \mathrm{d}s\leq \epsilon.
$
\end{lem}
\ifthenelse{\boolean{short}}{}{
\begin{proofof}{ \em Lemma \ref{lemma:2}.\ }
Let $\varepsilon>0$ and \( \cF \) be a function as in the assumptions of Lemma \ref{lemma:2}.
Let a parameter \( \delta > 0 \) that will be selected later.
    For each $n\in\NN$, on the interval \([x_n - \delta, x_n + \delta]\cap [0,\infty)\), define $\hat\cF$ as a {linear interpolation} between \( c_{n-1} \) and \( c_n \).
 Outside these intervals, set \( \hat\cF(x) = \cF(x) \). Now let us define the function $\tilde \cF:[0,\infty)\rightarrow \RR$ by 1) for all $x\in[x_n, x_n + \delta]$, $
        \tilde \cF(x) = c_{n-1} + \frac{x - x_n}{\delta}(c_n - c_{n-1})$; 2) for all $x\in[x_n - \delta, x_n]\cap[0,\infty)$, $
        \tilde \cF(x) = c_{n-1} + \frac{x - x_n + \delta}{\delta}(c_n - c_{n-1})$; 3) elsewhere, \( \tilde \cF(x) = \cF(x) \).

        Due to the third condition of the definition of $\tilde \cF$, it holds
        \begin{equation}\nonumber
        \begin{split}
        \int_0^\infty& \|\cF(x) - \tilde\cF(x)\|\ \mathrm{d}x\\
        &=
        \sum_{n\in\NN} \int_{\max(0,x_n-\delta)}^{x_n+\delta}\|\cF(x) - \tilde\cF(x)\|\ \mathrm{d}x
        \\
        &\leq
\delta         \sum_{n\in\NN} \|c_n-c_{n-1}\|
       \end{split}
       \end{equation}
where the last sum (and thus the next to last one too) is bounded because $\cF$ is of bounded variation. Selecting $\delta=\frac{\epsilon}{V(\cF,[0,\infty)}$, we get the conclusion of Lemma \ref{lemma:2}.
\end{proofof}}

\subsection{Proof of Theorem \ref{th:1}}
\label{sec:pro:th:1}
\begin{proofof}{ \em Theorem \ref{th:1}.\ }
Let us prove both parts of this theorem separately.
\newline
$\bullet$ Let $\cL$, $\cP$, $\cW_{i}$, $\csigma_{i}$, $i=1,\ldots, \ell$, $r$ and $\varepsilon$ as in the first part of Theorem \ref{th:1}.

Let $\mu>0$ that will be selected later. Applying Lemma \ref{lemma1} with $\nu=\frac{\epsilon}{r}$ and $\cF=\cL^\top$, there exists $0<\Delta \tau\leq 1$ and $L$ in $\RR^n$ such that,
for  all $j=0,\ldots, n$, and for all $\tau$ in $[j\Delta \tau,(j+1)\Delta \tau )$,
$$
\|L^{(:j)}-\cL( \tau)\|\leq
\frac{\mu}{r}\ .
$$
Therefore the first hidden state $Z_0$ of (\ref{12:juin.2}) and the first hidden state $\cZ_0$ of (\ref{7:juillet.1}) satisfy, for all $X$ in $\RR^p$, $\| X\| \leq r$, for all $j=0,\ldots, n$, for all $\tau \in [j\Delta \tau,(j+1)\Delta \tau )$,
\begin{equation}
\label{first:hidden:small}
|Z_0^{(j)}-\cZ_0( \tau)|\leq
\mu\ .
\end{equation}
Moreover, applying Lemma \ref{lemma1} to $\cW_{1}$ twice on the interval $[0,1)$, we get the existence of $n\in\NN\setminus\{0\}$ and a matrix $W\in \RR^{n\times n}$ such that, for all $j=0,\ldots, n$, for all $\tau \in [j\Delta \tau,(j+1)\Delta \tau )\cap[0,1)$, for all $i=0,\ldots, n$, for all $s \in [i\Delta \tau,(i+1)\Delta \tau )\cap[0,1)$,
\begin{equation}\label{first:approx}
|W_1^{(j,i)}-\cW_1( \tau,s)|\leq
\mu\ ,
\end{equation}
where $W_1^{(j,i)}$ is the $(j,i)$ entry of the the matrix $W_1$.
By uniform continuity of $\csigma_1$, we can select $0<\mu<1$ such that for all $\tau\in [j\Delta \tau,(j+1)\Delta \tau )$, and for all
$Z$, $Z^\prime$ in $\RR^n$ satisfying $\|Z- Z'\|\leq \mu$, it holds
\begin{equation}\label{eq:21:nov}
|\csigma_1(\tau,Z)-\csigma_1(\tau,Z^\prime)|\leq \epsilon\ .
\end{equation}

Let $\sigma_1:\RR^n\rightarrow \RR^n$ be the componentwise activation function, defined, for all $j=0,\ldots, n$, and for all $Z$ in $\RR^n$,
\begin{equation}\label{def:eq:sigma}
\sigma_1^{(j)}(Z)=\csigma_1(j\Delta\tau, Z^{(j)})
\ .\end{equation}
The function $\sigma_1$ is uniformly continuous on $[0,1)$ because the function $\csigma_1(j\Delta\tau,.)$ is also uniformly continuous, for all $j=0,\ldots, n$.

Let $M$ be a bound of the first hidden state $Z_0$ of (\ref{12:juin.2}) and of $\cW$. We get, for all $j=0,\ldots, n$, for all $\tau \in [j\Delta \tau,(j+1)\Delta \tau )\cap[0,1)$, for all $i=0,\ldots, n$, for all $s \in [i\Delta \tau,(i+1)\Delta \tau )\cap[0,1)$,
\begin{align}
\nonumber &|W_1^{(j,i)}Z_0^{(i)}-\cW_1( \tau,s)\cZ_0( s)|\\
\nonumber& \qquad  \qquad\leq
|W_1^{(j,i)}Z_0^{(i)}-\cW_1( \tau,s)Z_0^{(i)}|\\
\nonumber &\qquad  \qquad\phantom{\leq}+
|\cW_1( \tau,s)Z_0^{(i)}-\cW_1( \tau,s)\cZ_0( s)|\\\nonumber &
\qquad  \qquad\leq M ( |W_1^{(j,i)}-\cW_1( \tau,s)Z_0^{(i)}|+
|Z_0^{(i)}-\cZ_0( s)|)
\\\label{first:approx:1}
&\qquad  \qquad\leq 2M \mu
\end{align}
where
\eqref{first:hidden:small} and \eqref{first:approx} have been used in the last inequality.
Summing over $i$, we get, for all $j=0,\ldots, n$ and for all $\tau$ in $[j\Delta \tau, (j+1)\Delta \tau)\cap [0,1)$,
\begin{align*}
\sigma_1^{(j)}(\sum_{i=0,\ldots, n} W_1^{(:i)}Z_0^{(i)})\Delta \tau
= \csigma_1(j\Delta \tau,\sum_{i=0,\ldots, n} W_1^{(ji)}Z_0^{(i)})\Delta \tau
\end{align*}
where \eqref{def:eq:sigma} has been used. Therefore,
\begin{align*}
&| \sigma_1^{(j)}(\sum_{i=0,\ldots, n}W_1^{(:i)}Z_0^{(i)})\Delta \tau- \csigma_1(\tau,\int_0^1\cW_1( \tau,s)\cZ_0( s)\ \mathrm{d}s)|
\\
&\qquad \leq |\big(\csigma_1(j\Delta \tau,\sum_{i=0,\ldots, n} W_1^{(j,i)}Z_0^{(i)})\\
&\qquad \qquad  \qquad-\csigma_1(j\Delta \tau,\cW_1( \tau,i\Delta\tau)\cZ_0( i\Delta\tau))\big)\Delta \tau|
\\
&\qquad \phantom{\leq} + |\csigma_1(j\Delta\tau,\sum_{i=0,\ldots, n}\cW_1( \tau,i\Delta\tau)\cZ_0( i\Delta\tau))\Delta \tau\\
&\qquad \qquad- \csigma_1(\tau,\int_0^1\cW_1( \tau,s)\cZ_0( s)\ \mathrm{d}s)|
= T_1+T_2
\end{align*}
using the triangular inequality and for obvious definitions of the two terms $T_1$ and $T_2$.
Using equations \eqref{first:hidden:small}, \eqref{first:approx}, and \eqref{eq:21:nov}, it holds
$
T_1\leq (n+1)\Delta \tau\mu\leq 2\mu
$,
where $n\Delta \tau \leq 1$ and
$\Delta\tau\leq 1$ have been used.
Moreover, using the approximation of integrals by rectangular scheme (see e.g. \cite[Chapter 3]{StoerBulirsch02}) and up to decrease $\Delta\tau$, we have
$T_2\leq \epsilon$.
Therefore the second hidden state $Z_1$ of (\ref{12:juin.2}) and the second hidden state of (\ref{7:juillet.1}) satisfy, for all $X$ in $\RR^p$, $\| X\| \leq r$, for all $j=0,\ldots, n$, for all $\tau \in [j\Delta \tau,(j+1)\Delta \tau )$,
\begin{equation}
\label{second:hidden:small}
|Z_1^{(j)}-\cZ_1( \tau)|\leq
3 \epsilon\ .
\end{equation}
Proceeding similarly for the $l$ hidden layers until the hidden-to-output equation, we get \eqref{eq:discreti} (with $(2 M  +1)\epsilon$ instead of $\epsilon$).
This concludes the proof of the first part of Theorem \ref{th:1}.

$\bullet$
Let $(L^n)_{n>0}$, $(P^n)_{n>0}$, $(W^n_i)_{n>0}$, $(\sigma^n_i)_{n>0}$, for $i=1,\ldots, \ell$, $r$ and $\varepsilon$ be as in the second part of the theorem.

Let $\cL:[0,\infty)\rightarrow \RR^p$ be a piecewise constant function such that we have, for all $j\in\NN$, for all $n\geq j$,
$
\cL(j)=L^{n(j:)}
$, with discontinuities only in $\NN$.
Such a function could be built by a recurrence argument and exists because the sequence $(L^n)_{n>0}$ is consistent. Moreover, since the sequence $(L^n)_{n>0}$ is of bounded variation, then the function $\cL$ is of bounded variation. Now, using Lemma~\ref{lemma:2}, define
$\tilde \cL$ as the continuous function so that
$$
\int_0^\infty
\|\cL(s)-\tilde \cL(s)\|\ \mathrm{d}s\leq \epsilon\ .
$$
Therefore the first hidden state $Z_0$ of (\ref{12:juin.2}) and the first hidden state $\cZ_0$ of (\ref{7:juillet.1}) satisfy,
for all $X$ in $\RR^p$, $\| X\| \leq r$, for all $n$ and for all $j=0,\ldots, n$, for all $\tau \in [j\Delta \tau,(j+1)\Delta \tau )$,
$$
|Z_0^{n(j)}-\cZ_0( \tau)|\leq
\epsilon\ .
$$

Moreover define the function $\cW$ as the piecewise constant function such that for all $i,j\in\NN$, for all $n\geq j$,
$
\cW_1(i,j)=W_1^{n(i,j)}
$,
with discontinuity point only at $(i,j)$. The function $\cW_1$ is of bounded variation because the sequence $(W^n_1)_{n>0}$ is of bounded variation. Let $\widetilde \cW$ be a continuous function given by Lemma \ref{lemma:2} such that for all $n$, for all $i=0,\ldots, n$, $j=0,\ldots, n$, for all $s \in [i\Delta \tau,(i+1)\Delta \tau )$, for all $\tau \in [j\Delta \tau,(j+1)\Delta \tau )$, it holds
$$
\int_0^\infty |W_1^{n(i,j)}-\widetilde\cW_1( \tau,s)| \mathrm{d}s \leq
\frac{\epsilon}{r}\ ,
$$
and
$$
\int_0^\infty\int_0^\infty |W_1^{n(i,j)}-\widetilde\cW_1( \tau,s)| \mathrm{d}s \mathrm{d}\tau\leq
\epsilon\ .
$$
Similarly, define a continuous function $\widetilde\csigma_1:[0,\infty)\rightarrow \RR$ from a piecewise constant function  $\csigma_1$, built from the consistent sequence $(\sigma^n_i)_{n>0}$ of bounded variation, for $i=1,\ldots, \ell$, and using Lemma \ref{lemma:2}.
Therefore it holds, for all $X$ in $\RR^p$, $\| X\| \leq r$, for all $n$ and for all $j=0,\ldots, n$, for all $\tau \in [j\Delta \tau,(j+1)\Delta \tau )$,
$$
|Z_1^{n(j)}-\cZ_1( \tau)|\leq
\epsilon
$$
where $\cZ_1$ is given by
$
\cZ_{1}(\tau ) =
  \int_0^\infty \widetilde\csigma_{1} (\tau,\tilde  \cW _{1}(\tau,s ) \tilde  \cZ_0(s))  \ \mathrm{d}s  .
$
Now define a continuous function $\widetilde\cP:[0,\infty)\rightarrow \RR^q$ from the consistent sequence $(P^n){n>0}$ of bounded variation and using Lemma \ref{lemma:2}.
By doing so up to the last hidden layer of (\ref{12:juin.2}), we can build a DeepCNN as given by
\begin{equation}\label{7:juillet.1:infty}
\begin{split}
\cZ_0(\tau)&= \tilde \cL(\tau) X,  \quad \forall \tau\in [0,\infty), \\
\cZ_{i+1}(\tau ) & =
  \tilde  \csigma_{i+1} (s,\int_0^\infty\tilde  \cW _{i+1}(\tau,s )  \cZ_i(s)\ \mathrm{d}s )
  \ ,\\
  &\qquad \forall \tau\in [0,\infty),\; \forall i=0,\ldots, \ell-1, \\
Y&=  \int_0^\infty\tilde \cP (\tau )\cZ _\ell (\tau) \ \mathrm{d}\tau\ .
\end{split}
\end{equation}
Moreover, there exists a sufficiently large integer $\overline{n}$ such that, for all $X\in\RR^p$ satisfying $\|X\|\leq r$, the distance between the outputs of \eqref{12:juin.2} with $n\geq \overline{n}$ neurons and of (\ref{7:juillet.1:infty}) is less than $\varepsilon$:
$$
\forall n\geq \overline{n}, \|X\|\leq R \Rightarrow \| Y_{(\ref{7:juillet.1:infty})}- Y_{\eqref{12:juin.2}}\|\leq \varepsilon
\ .$$
The DeepCNN \eqref{7:juillet.1:infty} and \eqref{7:juillet.1} are equivalent up to a change of variable. More specifically, by using the change of variables $f:[0,1)\rightarrow [0,\infty)$ defined by $f(\tau') =\tan(\frac{\pi}{2}\tau')$, for all $\tau'\geq 0$ in the functions $\tilde\cL$, $\tilde \cP$, $\tilde\cW_{i}$, $\tilde\csigma_{i}$, $i=1,\ldots, \ell$, we may rescale all domains of definition and all integrals from $[0,\infty)$  in \eqref{7:juillet.1:infty} to $[0,1)$ as in \eqref{7:juillet.1}.
\newline
This concludes the proof of the second and last item of Theorem \ref{th:1}.
\end{proofof}

\subsection{Proof of Theorem \ref{th:2}}
\label{sec:pro:th:2}
\begin{proofof}{ \em Theorem \ref{th:2}.\ }Let us prove both parts of this theorem separately.

$\bullet$ Let $L$, $P$, $\cW$, $\csigma$, $r$ and $\varepsilon$ as in the start of Theorem~\ref{th:2}. Let $\mathcal{L}$ be a Lipchitz constant of the right-hand side of the second line of (\ref{7:juillet.3.2}) on the ball of $\RR^p$ with radius $r$.
\newline
Consider the time step $\frac{1}{\ell}$ for $\ell>0$ sufficiently large so that $\mathcal{L}<\ell$ and a partition $t_0=0<t_1<\cdots<t_\ell=T$ with step $\frac{1}{\ell}$. Then, by Euler discretization, for all input $X\in \RR^p$ satisfying $\|X\|\leq r$, it holds, for all $i=0,\ldots, \ell$,
\begin{equation}\label{4:dec}
\| Z_i-\cZ(t_i)\| \leq \frac{C}{\ell}
\end{equation}
where $C$ is a constant that does not depend on $\ell$. Now, up to enlarging $\ell$ so that $\max\{1,\|P\|\}C\frac{1}{\ell}\leq \epsilon$, it follows from \eqref{4:dec} with $i=\ell$ that
$$
\| Z_\ell-\cZ(T)\|\leq \frac{\epsilon}{\max\{1,\|P\|\}},
$$
and with the last line of (\ref{7:juillet.3.2}) and of \eqref{12:juin.3}, it holds
$
\| Y_{(\ref{7:juillet.3.2.2})}- Y_{\eqref{12:juin.3}}\|\leq \| P\| \| Z_\ell-\cZ(T)\|
\leq \varepsilon
$,
which concludes the proof of the first part of Theorem \ref{th:2}.

$\bullet$ Let
$L$, $P$, $((W^\ell_i)_{i=1,\ldots,\ell})_{\ell>0}$, $((\sigma^\ell_i)_{i=1,\ldots,\ell})_{\ell>0}$, $r$ and $\varepsilon$ as in the assumptions of the second part of Theorem \ref{th:2}.

For all $X\in \RR^n$ satisfying $\|X\| \leq r$, the sequence of hidden states $((Z_i^\ell)_{i =1,\ldots, \ell})_{l>0}$ given by \eqref{12:juin.3} is uniformly bounded and equicontinuous. Then we can define smooth
interpolating functions $(\cZ_\ell)_{\ell>0}:[0,T]\rightarrow \RR^n$ by, for all $i=0,\ldots, \ell-1$,
\begin{equation}\label{9:dec}
\cZ_\ell(\frac{i T}{\ell})=Z_i^\ell
\ ,
\end{equation}
and by an interpolation on
each interval $
[\frac{i}{\ell}T,\frac{i+1}{\ell}T]$, and such that $(\cZ_\ell)_{\ell>0}$ is uniformly bounded and equicontinuous. Moreover, due to \eqref{12:juin.3} and \eqref{9:dec}, it holds
\begin{eqnarray}
\nonumber
\!\!\frac{\ell}{T}
(\cZ_\ell(\frac{(i+1) T}{\ell})-\cZ_\ell(\frac{i T}{\ell}))&=&\frac{\ell}{T}
(Z_{i+1}^\ell-Z_i^\ell)\\\label{5:dec:2}
&=&\frac{\ell}{T}
\sigma_{i+1}^\ell(W_{i+1} ^\ell Z_i^\ell)
\end{eqnarray}
and since the sequence $((\ell\sigma_{i+1}^\ell(W_{i+1}^\ell Z_i^\ell ))_{i =1,\ldots, \ell})_{l>0}$ is equicontinuous and uniformly bounded,
by the Arzela-Ascoli theorem, there exists a subsequence of $(\cZ_\ell)_{\ell>0}$ that converges uniformly to continuously differentiable function $\cZ:[0,T]\to\mathbb{R}^n$. Similarly from the sequence of matrices $((W_i^\ell)_{i =1,\ldots, \ell})_{l>0}$, we can
define a matrix function sequence $(\cW_\ell)_{\ell>0}:[0,T]\rightarrow \RR^{n\times n}$ that is uniformly bounded and equicontinuous. Thus, there exists a subsequence that converges to a matrix function sequence $\cW:[0,T]\rightarrow \RR^{n\times n}$.

Similarly there exists a continuous function (not necessarily an activation function) $\csigma: \RR^n\rightarrow\RR^n$ and a subsequence of $(\frac{\ell}{T}(\sigma^\ell_i)_{i=1,\ldots,\ell})_{\ell>0}$ that converges to $\csigma$.
Due to \eqref{5:dec:2}, by uniform convergence of this subsequence of $(\frac{\ell}{T}(\sigma^\ell_i)_{i=1,\ldots,\ell})_{\ell>0}$, we have
$$
\begin{array}{c}
\dot \cZ (t)=\csigma(\cW( t) \cZ(t)) \ , \forall t\in[0,T]\ ,
\\
\cZ (0) = Z _0 \ , \; \cZ(T)=\lim_{\ell\rightarrow \infty}\cZ_\ell \ ,
\end{array}
$$
concluding the proof of the second part and the proof of Theorem \ref{th:2}.
\end{proofof}

\begin{rem}
    At the beginning of the proof of Theorem \ref{th:2}, an Euler discretization is used. Other numerical methods could be employed to achieve a better error order, such as the Runge–Kutta methods, see e.g., \cite[Chapter 7]{stoer1980introduction}.
\end{rem}

\section{Nomenclature}
\begin{table}[ht]
\centering
\caption{Nomenclature}
\begin{tabular}{ll}
\hline
\textbf{Acronym/Symbol} & \textbf{Meaning} \\
\hline
\multicolumn{2}{l}{\textit{Neural Network Architectures}} \\
CNN          & Continuous Neural Network\\
DeepNet      & Deep Neural Network\\
DeepCNN      & Deep Continuous Neural Network\\
DeepResNet   & Deep Residual Neural Network\\
DeepResCNN   & Deep Residual CNN\\
NeuralODE    & Neural Ordinary Differential Equation\\
NeuralResODE & Neural Residual ODE \\
NeuralCDE    & Neural Controlled Differential Equation\\
RNN          & Recurrent Neural Network\\
DiPaNet      & Distributed Parameter NN\\
DiPaResNet   & Distributed Parameter Residual NN\\
\hline
\multicolumn{2}{l}{\textit{Other Acronyms}} \\
NTK          & Neural Tangent Kernel\\
NIDE         & Neural Integro-Differential Equation\\
NN           & Neural Network\\
ODE          & Ordinary Differential Equation \\
PDE          & Partial Differential Equation \\
\hline
\end{tabular}
\label{tab:nomenclature}
\end{table}
}

\bibliography{cp_short}

\end{document}